\documentclass{article} % For LaTeX2e
\usepackage{collas2023_conference,times}
\usepackage{easyReview}

\usepackage{amsmath,amsfonts,bm}

\def\eqref#1{equation~\ref{#1}}
\def\1{\bm{1}}

\DeclareMathAlphabet{\mathsfit}{\encodingdefault}{\sfdefault}{m}{sl}
\SetMathAlphabet{\mathsfit}{bold}{\encodingdefault}{\sfdefault}{bx}{n}

\usepackage{hyperref}
\hypersetup{
    colorlinks=true,
    linkcolor=red,
    filecolor=magenta,
    urlcolor=blue,
    citecolor=purple,
    pdftitle={Overleaf Example},
    pdfpagemode=FullScreen,
    }

\title{SF-FSDA: Source-Free Few-Shot Domain Adaptive Object Detection with Efficient Labeled Data Factory}

\author{Han Sun \\
EPFL\\
Switzerland \\
\texttt{han.sun@epfl.ch} \\
\And % Use And to have authors side by side
Rui Gong \\
ETH Zurich \\
Switzerland \\
\texttt{gongr@vision.ee.ethz.ch}
\And
Konrad Schindler\\
ETH Zurich \\
Switzerland \\
\texttt{schindler@ethz.ch}
\And
Luc Van Gool\\
ETH Zurich \\
Switzerland \\
\texttt{vangool@vision.ee.ethz.ch}
}

\preprintcopy % Uncomment for the preprint version, but NOT for submission.
\usepackage{etoolbox}
\newcommand{\MyMapTemplatePrefixc}[4]{\expandafter#1\csname#3#4\endcsname{#2{#4}}} % it remembles a template: \#3#4 --> #2{#4}
\forcsvlist{\MyMapTemplatePrefixc {\def} {\mathcal}{c}} {A,B,C,D,E,F,G,H,I,J,K,L,M,N,O,P,Q,R,S,T,U,V,W,X,Y,Z}  % e.g., \cA --> \mathcal{A}

\newcommand{\MyMapTemplateNoPrefix}[3]{\expandafter#1\csname#3\endcsname{#2{#3}}}
\forcsvlist{\MyMapTemplateNoPrefix {\def} {\mathbf} } {0,1,a,b,c,d,e, f, g, h, i, j, k, l, m, n, o, p, q, r, s, t, u, v, w, x, y, z} % e.g., \a --> \mathbf{a}
\forcsvlist{\MyMapTemplateNoPrefix {\def} {\mathbf} } {A,B,C,D,E,F,G,H,I,J,K,L,M,N,O,P,Q,R,S,T,U,V,W,X,Y,Z}  % e.g., \A --> \mathcal{A}

\makeatletter
\newcommand{\thickhline}{%
    \noalign {\ifnum 0=`}\fi \hrule height 1pt
    \futurelet \reserved@a \@xhline
}

\usepackage{multirow}
\usepackage{colortbl}
\definecolor{mygray}{gray}{.9}

\usepackage{subcaption}
\usepackage{color}
\usepackage{soul}
\usepackage{tcolorbox}

\begin{document}

\maketitle
\begin{abstract}
Domain adaptive object detection aims to leverage the knowledge learned from a labeled source domain to improve the performance on an unlabeled target domain. Prior works typically require the access to the source domain data for adaptation, and the availability of sufficient data on the target domain. However, these assumptions may not hold due to data privacy and rare data collection. In this paper, we propose and investigate a more practical and challenging domain adaptive object detection problem under both \emph{source-free} and \emph{few-shot} conditions, named as SF-FSDA. To overcome this problem, we develop an efficient labeled data factory based approach. Without accessing the source domain, the data factory renders i) infinite amount of synthesized target-domain like images, under the guidance of the few-shot image samples and text description from the target domain; ii) corresponding bounding box and category annotations, only demanding minimum human effort, \ie, a few manually labeled examples. On the one hand, the synthesized images mitigate the knowledge insufficiency brought by the few-shot condition. On the other hand, compared to the popular pseudo-label technique, the generated annotations from data factory not only get rid of the reliance on the source pretrained object detection model, but also alleviate the unavoidably pseudo-label noise due to domain shift and source-free condition. The generated dataset is further utilized to adapt the source pretrained object detection model, realizing the robust object detection under SF-FSDA. The experiments on different settings showcase that our proposed approach outperforms other state-of-the-art methods on SF-FSDA problem. Our codes and models will be made publicly available.
\end{abstract}

\section{Introduction}\
Object detection, which aims at recognizing and localizing the object instances of certain classes in an image, is a fundamental problem in computer vision. Driven by the rapid development of deep learning and the availability of large-scale datasets, object detection has achieved great advancement over the past decade~\cite{ren2015faster,liu2016ssd,redmon2016you,carion2020end}. However, the performance and generalization ability of the detection system is highly dependent on the availability of manually labeled and diverse datasets, whose labor cost for annotation can be extremely expensive. When applied to the images of a different distribution with training images, the detection models typically exhibit poor generalization, which is common in real applications due to the difference in weather, illumination, object appearance, \etc. Thus, recently, domain adaptive object detection problem has been studied~\cite{chen2018domain,saito2019strong,hsu2020every,khodabandeh2019robust,vs2021mega,ramamonjison2021simrod}, which aims to transfer the knowledge learned from the labeled source domain to the unlabeled target domain to train the robust cross-domain object detection model, reducing the effort and cost of human annotation for the target domain. 

\begin{figure*}[t]
    \centering
    \includegraphics[width=1\linewidth]{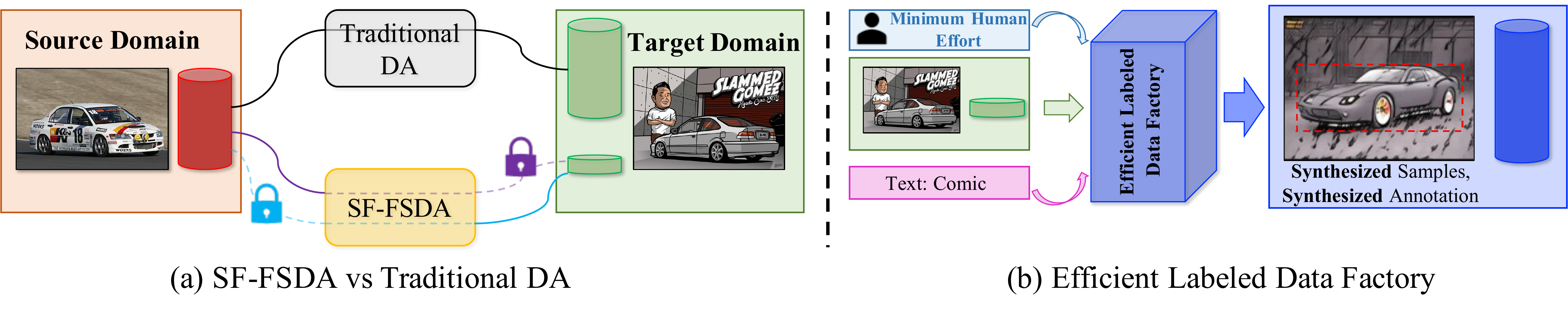}
    \caption{(a) Traditional domain adaptive object detection (DA) problem \emph{vs}\onedot our proposed SF-FSDA problem. Our SF-FSDA problem considers the \emph{few-shot} and \emph{source-free} conditions. (b) Our efficient labeled data factory can synthesize abundant target domain like image samples and corresponding bounding box and category annotation automatically, by providing the few-shot samples guidance, text guidance, and minimum human effort (\ie, few-shot manual label). The efficient labeled data factory is initialized with publicly available pretrained GAN model weights, and does not rely on the access to the source domain images and/or other public image dataset. The height of the cylinder represents the number of image samples.}
    \label{fig:problem_overview}
\end{figure*}

Generally, existing domain adaptive object detection works reduce the domain shift between the source domain and the target domain, by matching and aligning the source and target domain representations in some space (input space~\cite{hoffman2018cycada,inoue2018cross,bhattacharjee2020dunit} and/or feature space~\cite{chen2018domain,deng2021unbiased}) through the typical techniques of adversarial learning~\cite{rezaeianaran2021seeking,saito2019strong}, pseudo-label~\cite{roychowdhury2019automatic,kim2019self,munir2021ssal}, and image translation~\cite{hsu2020progressive,inoue2018cross}. They typically assume that, i) the source domain images are accessible when adapting to the target domain, and/or ii) there are abundant images available in the target domain. However, both of these assumptions may not hold in real applications. For example, the data privacy rules and the limited data transmission capacity can break the assumption i), \ie, inducing the \emph{source-free} condition, while the rare species image collection and the special medical applications can hinder the assumption ii), \ie, causing the \emph{few-shot} condition. Even though some more recent works touch the domain adaptive object detection problem under the source-free condition~\cite{li2020free} or the few-shot condition~\cite{wang2019few}, none of them consider the problem when both the source-free and few-shot conditions exist at the same time. Instead, this work studies the domain adaptive object detection problem under both \emph{source-free} and \emph{few-shot} conditions, named as SF-FSDA, \ie, the source domain images are not accessible when adapting the object detection model to the target domain, and there are only a few samples available in the target domain (see Fig.~\ref{fig:problem_overview}\textcolor{red}{a}).

The current available domain adaptive object detection approaches can tackle isolated one of the two conditions in SF-FSDA, but cannot deal with the two conditions at the same time. More specifically, pseudo-label based techniques are popularly utilized in source-free conditions~\cite{li2020free,kundu2021generalize}, but are not capable of handling the few-shot condition, since it relies on enough samples to reduce the pseudo-supervision noise brought by the domain gap. In contrast, adversarial learning~\cite{motiian2017few,ganin2015unsupervised} and image translation~\cite{luo2020adversarial,ma2018exemplar} based methods can operate under few-shot conditions, but require access to the source domain. 

In order to address the challenging SF-FSDA problem, we propose an efficient labeled data factory based method (see Fig.~\ref{fig:problem_overview}\textcolor{red}{b}), which i) synthesizes abundant target domain like images guided by the few-shot samples and the text description from the target domain, without accessing the source domain image; and ii) automatically generates the corresponding object bounding box and category annotations, with the help of minimum human effort, \ie, few-shot manual annotation. Compared to the existing image translation based approach~\cite{zhu2017unpaired,park2020contrastive,huang2018multimodal,liu2019few,isola2017image}, our proposed data factory based method does not require the availability to the source domain (\emph{source-free condition}), and effectively exploits the few-shot image (\emph{few-shot condition}) and text knowledge from the target domain for the image synthesis. The text knowledge provides the general guidance, \eg, what the ``comic" style images look like, while the few-shot images offer the specific guidance, \eg, how the ``comic" images on the target domain are like. The text knowledge can also prevent the overfitting effect brought by the few-shot images. Thus, the text and few-shot images guidances promote each other to synthesize the more target-domain like images. Besides, the data factory based method renders the object bounding box and category annotation together with and guided by the image synthesis process, only requiring the few-shot manual annotation. Our data factory model is initialized with publicly available GAN pretrained weights, which are irrelevant to both the source and target domain. In this way, the efficient labeled data factory synthesizes the target domain on both the \emph{image level} and \emph{label level}. Alternatively, applying the source domain trained object detection model on our synthesized image can generate the pseudo-label, which however is noisy and low-quality due to the domain gap. Instead, our efficient labeled data factory can synthesize the detection label without relying on the source trained object detection model, generating higher quality annotation and easing the downstream domain adaptive object detection. Then the source domain trained object detection model is fine-tuned on our synthesized images and annotations to train the final object detection model.

In a nutshell, the key contributions of this paper are three-fold. \textbf{(1)} We propose the domain-adaptive object detection problem under the \emph{source-free} and \emph{few-shot} conditions, SF-FSDA, where there are only a few samples available on the target domain, and only the source-pretrained model is accessible for the adaptation to the target domain. \textbf{(2)} We develop the efficient labeled data factory based approach to SF-FSDA problem, where the efficient labeled data factory can automatically synthesize a number of the target domain like images and corresponding object detection labels, providing the text guidance, few-shot sample guidance and minimum human annotation. \textbf{(3)} The experimental results on different benchmarks prove the effectiveness of the proposed method for the SF-FSDA problem, serving as the strong baseline for further research. 

\section{Related Work}
\subsection{Image Synthesis.}
Recently, generative adversarial networks (GANs)~\cite{goodfellow2014generative} have become an active research area and boosted numerous applications, especially image synthesis. 
It is demonstrated that, given proper training, GANs can synthesize semantically meaningful data from standard data distributions. The current state-of-the-art GAN models~\cite{brock2018large,gao2019progan, zhang2018stackgan++} are able to generate high-quality realistic images of diverse categories. Recent style-based generators~\cite{karras2019style,karras2020analyzing,karras2020training,karras2021alias} produce impressive results and allow for style control via mapping noise vectors to a higher-dimensional semantic space, which inspires several extensions such as image manipulation~\cite{patashnik2021styleclip,gal2021stylegannada,zhu2021mind}, image editing~\cite{ling2021editgan,bau2021paint}, and dataset synthesis~\cite{zhang2021datasetgan}. 
Besides, various types of potential guidance~\cite{li2020manigan,ling2021editgan,dhamo2020semantic,collins2020editing,Patashnik_2021_ICCV} are utilized for controlling the image synthesis process, among which the most relevant to our work are text guidance and few-shot image guidance. The text guidance is typically provided by the large-scale vision-language pretrained model~\cite{radford2021learning}, by mapping the images and the text to the joint embedding space. More recent works~\cite{patashnik2021styleclip,gal2021stylegannada} then introduce the image and text consistency in the embedding space to regularize image synthesis. The few-shot image guided image synthesis works~\cite{saito2020coco,liu2019few,ojha2021few} aim to prevent image generation overfitting when only limited image samples are available. Different from the aforementioned works, our proposed data factory exploits both the text guidance and the few-shot images guidance together, promoting each other to further improve image synthesis in the target-domain (see ablation study in Sec.~\ref{sec:experiment_results}). The label synthesis of our data factory is related to the work of ~\cite{zhang2021datasetgan}, which, however, only focuses on the dense prediction task, \eg, semantic segmentation, and does not consider the domain adaptation problem. Instead, our proposed data factory tackles the domain adaptation problem with the few-shot samples and text guidance, and investigates the synthesis of object detection annotations.

\subsection{Domain Adaptive Object Detection.} Domain adaptation aims to transfer knowledge between the label-rich source domain and the unlabeled target domain to train the model that performs well on the target domain. In the past decades, it has been explored in different tasks, \eg, image classification~\cite{tzeng2017adversarial,gong2012geodesic,ganin2015unsupervised}, semantic segmentation~\cite{Tsai_adaptseg_2018,vu2018advent,tranheden2021dacs}, and object detection~\cite{chen2018domain,khodabandeh2019robust,vs2021mega}. Among the quite vast scope, the most relevant category to our work is domain adaptive object detection, where adversarial learning, image translation, and pseudo-label based methods are typically proposed and studied. Recently, considering more practical scenarios, some works explore the source-free~\cite{li2020free} or few-shot~\cite{wang2019few} domain adaptive object detection problem, respectively. More specifically, \cite{li2020free} tackles the source-free~\cite{kundu2020universal,yang2021generalized,kundu_cvpr_2020} domain adaptive object detection problem with the pseudo-label based technique. And \cite{wang2019few} studies the few-shot~\cite{motiian2017few} domain adaptive object detection problem through adversarial learning based method. However, none of the aforementioned works investigate both the \emph{source-free} and \emph{few-shot} conditions at the same time. In contrast, our SF-FSDA problem touches both \emph{source-free} and \emph{few-shot} conditions simultaneously, which is more challenging and practical. From the method aspect, instead of exploiting pseudo-label or adversarial learning, we synthesize the target domain-like images and the corresponding bounding box and category annotations together with the efficient labeled data factory, without accessing the source domain.

\subsection{Domain Transfer with Auxiliary Knowledge.} In some domain transfer related works, \eg, domain adaptation, domain generalization and domain randomization, the auxiliary knowledge from the public dataset is utilized as the bridge to connect the source domain and the target domain. For example, since the target domain image is not available for training, \cite{yue2019domain} randomizes the style of the source domain images utilizing the images from the public dataset ImageNet~\cite{deng2009imagenet}, to improve the generalization ability of the semantic segmentation model trained on the source domain. \cite{wu2021domain} adopts the auxiliary images from ImageNet to regularize the image classification model training in the adaptation process, to prevent the model from forgetting. However, these works all require access to the auxiliary images, which might not be practical due to data privacy regulations and data transmission capacity. Instead, our efficient labeled data factory takes the publicly available GAN pretrained weights~\cite{karras2020analyzing} as the auxiliary knowledge, which is more flexible and renders unlimited and unified image and label synthesis. 

\section{Method}
\subsection{Problem Statement}
For the problem of domain-adaptive object detection, we are given the labeled source domain $\cS=\{\x_s^i, \y_s^i\}_{i=1}^{N_s}$ and the unlabeled target domain $\cT=\{\x_t^i\}_{i=1}^{N_t}$, where $\x_s^i, \y_s^i$ represent the $i$-th image and the corresponding bounding box and category annotations for object detection in the source domain, and $\x_t^i$ denotes the $i$-th unlabeled image in the target domain. $N_s, N_t$ are the number of images in the source and target domain, respectively. Different from traditional domain adaptive object detection problem, we tackle the \emph{source-free} and \emph{few-shot} target conditions, \ie, $N_s\gg N_t$ and $\{\x_s^i, \y_s^i\}_{i=1}^{N_s}$ is not accessible during the adaptation process to $\{\x_t^i\}_{i=1}^{N_t}$, named SF-FSDA problem.

\noindent\textbf{Technical Challenges.} Compared to the traditional domain adaptive object detection problem, our proposed SF-FSDA problem introduces more challenging \emph{source-free} and \emph{few-shot} conditions. Previous techniques for domain adaptive object detection highly rely on adversarial feature learning~\cite{chen2018domain}, image-to-image translation~\cite{inoue2018cross}, and pseudo-label based self-training~\cite{roychowdhury2019automatic}. On the one hand, the challenge brought by \emph{source-free} condition is that, the previous adversarial feature learning and image-to-image translation based techniques require access to source data during the adaptation process to align the distribution between the source and target domains, making them not equipped to be engaged in our \emph{source-free} setting. On the other hand, the challenge induced by \emph{few-shot} condition is that, the pseudo-label based self-training technique always relies on the availability of abundant target domain images to reduce the prediction noise and improve the prediction confidence on the target domain, which are difficult to operate in our \emph{few-shot} setting. Thus, both the \emph{source-free} and \emph{few-shot} conditions hinder the knowledge transfer between the source and target domains for object detection.

\noindent\textbf{Motivation.} As discussed in the aforementioned technical challenges, the \emph{source-free} and \emph{few-shot} conditions make it difficult to adapt guided by the object detection task. Thus, we aim to firstly adapt on the image level, \ie, synthesize the target domain like image. However, different from the previous image translation methods that rely on the access to the source domain and the target domain at the same time, the adaptation, guided by the few-shot samples and text from the target domain, on the publicly available trained GAN model provides more flexibility without accessing the source domain data. Moreover, in order to provide reliable guidance for the downstream object detection task, the method for synthesizing the corresponding object detection label is developed. Inspired by the observation that the trained GAN model encodes the rich knowledge related to the object category and position implicitly in the latent feature space, we introduce the label synthesis branch to produce the object category and bounding box annotation automatically, providing only minimum human effort, \ie, few-shot manual annotation. 

\subsection{Efficient Labeled Data Factory for SF-FSDA Problem}\label{sec:fact_method}
\begin{figure*}[t]
    \centering
    \includegraphics[width=\textwidth]{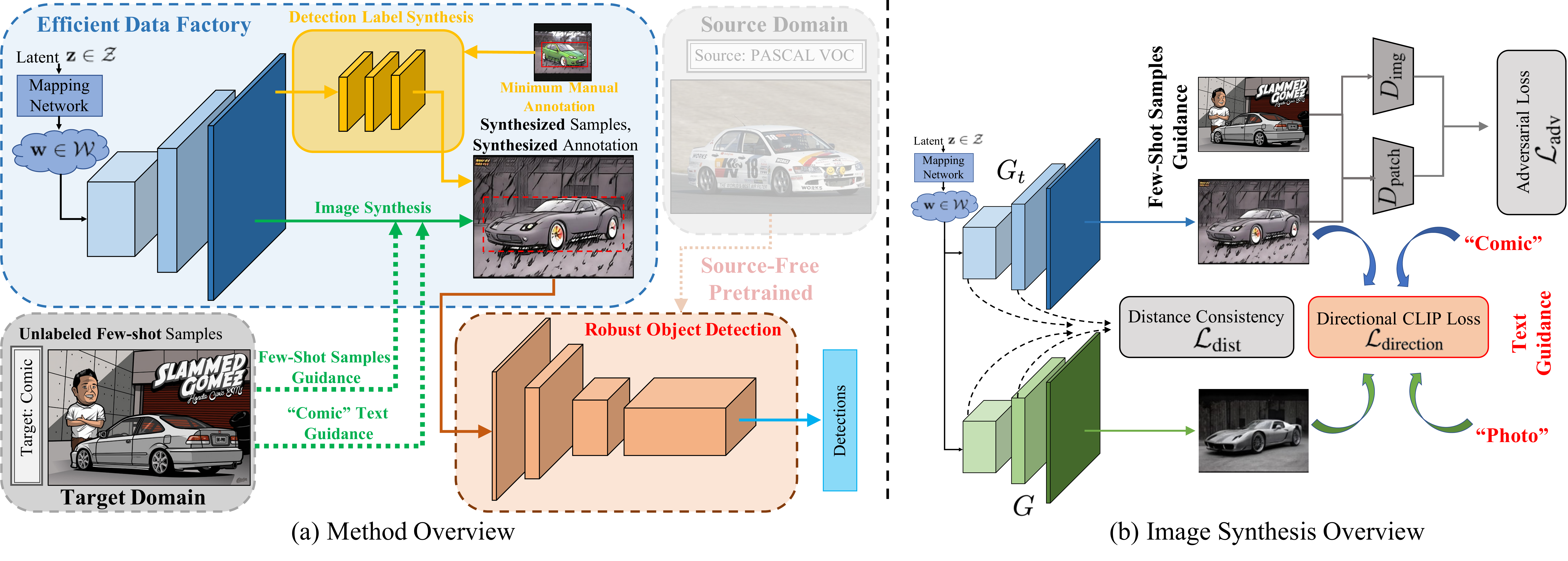}
    % \vspace*{-5mm}
    \vspace{-20pt}
    \caption{(a) Method overview. The well labeled PASCAL VOC dataset is taken as the source domain, while there are only few-shot unlabeled Comic samples available in the target domain. The aim is to train the domain-adaptive object detection model under the \emph{source-free} and \emph{few-shot} condition, \ie, only the source pretrained model and few-shot target domain samples are available for the adaptation to the target domain. The efficient data factory module is composed of the image synthesis branch and the detection label synthesis branch. The image synthesis branch is guided by the few-shot samples and text from the target domain, to synthesize the target domain-like image. The detection label synthesis branch aims to automatically synthesize the bounding box and category label, with the help of minimum human effort, \ie, few-shot manual annotation. (b) Image synthesis overview. The image synthesis branch is driven by the few-shot image samples guidance and the text guidance.
    }
    % \vspace{-15pt}
    \label{fig:method_overview}
\end{figure*}

In order to deal with the SF-FSDA problem, we propose the efficient labeled data factory based method, as shown in Fig.~\ref{fig:method_overview}. Since the SF-FSDA problem touches the \emph{source-free} setting, the whole training stage will be divided into, i) source-pretraining stage, ii) image and label synthesis stage and iii) target-adaptation stage. In the i) source-pretraining stage, the object detection model is trained on the source domain. Then in the ii) image and label synthesis stage, the efficient labeled data factory is driven by the few-shot image sample and text guidance from the target domain to synthesize the image with the image synthesis branch, and the label synthesis branch automatically synthesizes the corresponding object detection label by only providing the few-shot manual annotation. In the iii) target-adaptation stage, the synthesized image and corresponding label synthesized in stage ii) are exploited to fine-tune the source pretrained object detection model in stage i).

\noindent\textbf{Image Synthesis with Few-Shot Image Guidance.}
Given a publicly pretrained GAN model with the generator $G$, we aim to learn an adapted generator $G_{t}$ guided by the few-shot image samples $\{\x_t^i\}_{i=1}^{N_t}$ from the target domain $\cT$.
Following~\cite{ojha2021few}, the distance consistency regularization, $\cL_{dist}$, is utilized to preserve the original content and diversity of the image samples, and the anchor-based relaxed realism is adopted to further prevent the overfitting to the few-shot image samples. In more detail, the distance consistency regularization samples a batch of $N + 1$ noise vectors and uses their pairwise similarities in feature space to construct an $N$-way probability distributions for each image. The $N$-way probability of the image generated by the i-th noise vector is given by:
\begin{equation}
\begin{gathered}
y_i^{m}=\operatorname{Softmax}\left(\left\{\operatorname{sim}\left(G^m\left(z_i\right), G^m\left(z_j\right)\right)\right\}_{\forall i \neq j}\right) \\
y_i^{t, m}=\operatorname{Softmax}\left(\left\{\operatorname{sim}\left(G_{t}^m\left(z_i\right), G_t^m\left(z_j\right)\right)\right\}_{\forall i \neq j}\right),
\end{gathered}
\end{equation}
where similarity denotes the cosine similarity of activations at the $m^{th}$ layer between generator $G$ and $G_t$. The probability distributions converted from the similarities of the adapted model and the given publicly pretrained model are encouraged to be uniform by computing KL-divergence across the intermediate layers. With the anchor-based relaxed realism, a dual-discriminator training approach is deployed to prevent overfitting to the few-shot samples. An anchor region is defined as a subset of the entire input latent space $\cZ$. When sampled from these regions, we use a full image discriminator $D_{img}$. Outside of them, we enforce adversarial loss using a patch-level discriminator $D_{patch}$ to avoid overfitting to the few-shot samples.

We observe minor collapse in style with the training strategy in~\cite{ojha2021few} under our setting, \ie, the same color or pattern before successful adaptation. Different from~\cite{ojha2021few}, we relax the distance consistency regularization during different training phrases to allow a reasonable extent of object shape adaptation while still keeping the original image content. 
In the initial training phases, we compute the distance consistency only of the deep layers (\ie, after 6th layer) of the generator. After training for certain epochs, we adapt the training strategy and only compare consistency on the shallow layers (\ie, before 10th layer) for detailed style adaptation and to preserve the content.

The objective of image synthesis with few-shot image guidance is,
\begin{equation}
\begin{aligned}
G_{t}^{*}=\arg \min _{G_{t}} \max _{D_{\text {img }}, D_{\text {patch }}} & \mathcal{L}_{\text {adv }}\left(G_{t}, D_{\text {img }}, D_{\text {patch }}\right) +\lambda_1 \mathcal{L}_{\text {dist }}\left(G_{t}, G\right),
\end{aligned}
\label{eq:img_guid}
\end{equation}
where $\cL_\text{adv}$ represents the adversarial loss, and $\lambda_1$ is the hyper-parameter to balance the adversarial loss and the distance consistency regularization loss. 

\noindent\textbf{Image Synthesis with Text guidance.}
Besides the few-shot image samples from the target domain, the text description about the target domain is available with no effort required, \eg ``cartoon" and ``watercolor." To fully exploit and transfer the knowledge from the target domain to imitate its distribution, text guidance from the target domain can be leveraged to guide the image synthesis of the data factory with the help of contrastive language-image pre-training (CLIP) models~\cite{radford2021learning}. The main idea is to train the GAN model to make the generated images shift along the direction of the textually-described path in the CLIP embedding space~\cite{gal2021stylegannada}. Original and target texts are both self-defined to provide the desired shifting guidance. In order to obtain the image shifting direction during training, a dual-generator strategy is also deployed. We fix the pretrained generator $G$ to keep generating original images for comparison while optimizing the target generator $G_{t}$. Then the changing directions of text guidance and images can be expressed by,
\begin{equation}
\begin{gathered}
\Delta T=E_{text}\left(T_{\text {target }}\right)-E_{text}\left(T\right) \\
\Delta I=E_{img}\left(G_{t}(\z)\right)-E_{img}\left(G(\z)\right), 
\end{gathered}
\end{equation}
where $E_{text}$ and $E_{img}$ denote CLIP text and image encoders, respectively. $T$ and $T_{\text{target}}$ represent the text description of the pretrained GAN model and the target domain, \eg, ``photo" and ``comic." $\z$ is the input noise variable, \ie, $\z\in \cZ$. The directional loss introduced by text guidance can thus be described as,
\begin{equation}
\begin{gathered}
\mathcal{L}_{\text {direction}}(G, G_t, T, T_\text{target})=1-\frac{\Delta I \cdot \Delta T}{|\Delta I||\Delta T|}.
\end{gathered}
\end{equation}

Combined with the few-shot image guidance training objective in Eq.(\ref{eq:img_guid}), our final training objective with both the few-shot image guidance and the text guidance can be derived as,
\begin{eqnarray}
\begin{aligned}
G_{t}^{*}=\arg \min _{G_{t}} \max _{D_{\text {img }}, D_{\text {patch }}} & \mathcal{L}_{\text {adv }}\left(G_{t}, D_{\text {img }}, D_{\text {patch }}\right) \\
&+\lambda_{1} \mathcal{L}_{\text {dist }}\left(G_{t}, G\right) \\
&+\lambda_{2} \mathcal{L}_{\text {direction }}\left(G_{t}, G, T, T_{\text{target}}\right),
\end{aligned}
\label{eq:img_syn}
\hspace{-30pt}
\end{eqnarray}
where $\lambda_2$ is the hyper-parameter to balance the text guidance and other terms.

\noindent\textbf{Image Synthesis Training Strategy.} \label{sec:train_strg}
In order to further prevent the model from overfitting the few-shot image samples, we adopt the freezing strategy during the training. More specifically, the shallow layers of the original generator are frozen, while shallow layers of the discriminator are also frozen accordingly to further ensure a stable training process. The training strategy is simple yet effective in preventing overfitting under few-shot conditions.

\noindent\textbf{Label Synthesis.}
We can now get unlimited target-like samples with a successfully adapted image synthesis branch.
Previous research has proved that StyleGAN2~\cite{karras2020analyzing} learns a well-disentangled semantic latent space, where each channel controls some meaningful properties at different scales.
Intuitively, feature maps generated by those channels should be semantically informative enough to act as extracted features for downstream tasks, e.g., segmentation and detection.
Based on this assumption, we develop our object detection branch and get our training data with the following procedure: We sample a set of latent codes $\{z_{i}\}_{i=1}^{N_a}$ and generate their corresponding images $\{G_{t}(z_{i})\}_{i=1}^{N_a}$. Here $N_a$ denotes the number of manual annotations required to train the object detection task. Then we manually annotate these samples as our training data. During the training process, we deploy the generator $G_{t}$ as our backbone network and concatenate the intermediate convolutional feature maps generated by the latent codes as our encoded features for the matching images. 
A prediction head is built on these extracted features and trained for the object detection task.

Inspired by~\cite{zhou2019objects}, we use keypoint representations where each object is represented by its center point and the size of its bounding box. To detect objects presented in a synthesized image $\bar{\x}_t \in R^{W \times H \times 3}$, our goal is to predict a downsampled keypoint heatmap $\hat{\y} \in[0,1]^{{\frac{W}{r}} \times \frac{H}{r} \times C}$. $C$ denotes the number of classes for the prediction task, $r$ represents the downsampling stride, and $W, H$ are the width and height of the image. A prediction $\hat{\y}_{x, y, c} = 1$ represents a detected keypoint of class $c$, while $\hat{\y}_{x, y, c} = 0$ means background. For loss propagation, ground truth heatmap $\y$ is generated by converting each ground truth keypoint $p \in \mathcal{R}^{2}$ to its low-resolution equivalent $\tilde{p}=\left\lfloor\frac{p}{r}\right\rfloor$ and splatting those points using a Gaussian Kernel. The training loss is defined as a variant of focal loss~\cite{lin2018focal},
\begin{eqnarray}
\cL_{k}\!\!=\!\!\frac{-1}{N} \sum_{x y c}\left\{\begin{array}{cl}
\!\!\!\!\!\!\!\!\!\!\!\!\!\!\!\!\!\!\!\!\!\!\!\!\!\!\!\!\!\!\!\!\!\!\left(1-\hat{\y}_{x y c}\right)^{\alpha} \log \left(\hat{\y}_{x y c}\right) &\!\!\!\!\!\text { if } \y_{x y c}=1 \\
\!\!\!\!\left(1-\y_{x y c}\right)^{\beta}\left(\hat{\y}_{x y c}\right)^{\alpha} \log \left(1-\hat{\y}_{x y c}\right) &\! \text { otherwise, }
\end{array}\right.
\label{eq:lbl_k}
\hspace{-20pt}
\end{eqnarray}
where $\alpha$ and $\beta$ are hyper-parameters of the focal loss, while $N$ is the number of keypoints in image $\bar{\x}_t$ for normalization.

A local offset $\hat{\mathbf{o}} \in \mathcal{R}^{\frac{W}{r} \times \frac{H}{r} \times 2}$ is predicted and shared among all classes to recover the precise center point locations in compensation for the error caused by downsampling. The sizes of bounding boxes $\hat{\s} \in \mathcal{R}^{\frac{W}{r} \times \frac{H}{r} \times 2}$ of each class $c$ are regressed around the predicted center points, using a single shared prediction as well. Offset loss is computed only at locations of predicted keypoints $\tilde{p}$, while size loss is computed for each detected object $k$ with its predicted size $\hat{\s}_{p_{k}}$ around the center point $p_{k}$ and the ground truth bounding box size $\s_{k}$. Both keypoint offset and size predictions are trained with L1 loss,
\begin{equation}
\begin{gathered}
L_{off}=\frac{1}{N} \sum_{p}\left|\hat{\mathbf{o}}_{\tilde{p}}-\left(\frac{p}{r}-\tilde{p}\right)\right| \\
L_{s i z e}=\frac{1}{N} \sum_{k=1}^{N}\left|\hat{\s}_{p_{k}}-\s_{k}\right|.
\end{gathered}
\end{equation}

A two-layer convolutional head is built as the label synthesis branch for predicting $\hat{\y}$, $\hat{\mathbf{o}}$, and $\hat{\s}$ each and trained with a weighted sum of loss terms for these tasks,
\begin{equation}
\cL_{det}=\cL_{k} +\lambda_{off} \cL_{off} + \lambda_{size} \cL_{size},
\label{eq:lbl_syn}
\end{equation}
where $\lambda_{off}$ and $\lambda_{size}$ represent the hyper-parameters to balance the offset, size and keypoint prediction training loss.

\section{Experiment}
\subsection{Experimental Setup}
\noindent\textbf{In-Domain Experiments.} In order to verify the validity of our proposed efficient labeled data factory for automatically producing the images and corresponding object category and bounding box labels, we conduct the in-domain experiments, where our efficient labeled data factory is used to generate the images and object bounding box and category annotations, \emph{without domain adaptation}. Then, the generated images and labels are exploited to train the object detection model, to recognize the object instances on the same or similar domain, \emph{i.e.,} in-domain object detection. More specifically, in our experiment, the object detection model is trained on the natural images and annotations (see Fig.~\ref{fig:visual_comp}\textcolor{red}{c}) synthesized by our data factory, and tested on the PASCAL VOC datset.

\noindent\textbf{SF-FSDA Cross-Domain Experiments.} For the purpose of proving the helpfulness of our proposed efficient labeled data factory for domain adaptation, the SF-FSDA cross-domain experiments are explored, where the data factory is trained with the guidance of the \emph{text} and/or the \emph{few-shot samples} from the target domain. Furthermore, the synthesized images and labels are utilized to fine-tune the source domain pretrained object detection model, to adapt the model to the target domain, \emph{i.e.,} SF-FSDA cross-domain object detection. In our experiments, we aim at realizing SF-FSDA, under PASCAL VOC (source) $\rightarrow$ Clipart and Comic (target), respectively. 

\noindent\textbf{Dataset.} \emph{PASCAL VOC}: PASCAL VOC 2007 \& 2012 datasets~\cite{everingham2010pascal} contain natural objects with manually labeled bounding box and category annotations. In the in-domain experiments, the test set with cat and car objects is utilized to evaluate the performance of the object detection model. In the SF-FSDA cross-domain experiments, the training set including labeled car and cat images is taken as the source domain for training. 
\emph{Clipart1k}: Clipart1k dataset~\cite{inoue2018cross} covers clipart images, exhibiting a large domain shift compared to PASCAL VOC dataset. In the SF-FSDA cross-domain experiments, 12 unlabeled images are exploited as the few-shot target samples for training, and the test set containing cat and car objects is taken for the model performance evaluation on SF-FSDA.
\emph{Comic2k}: Comic2k dataset~\cite{inoue2018cross} consists of comic images, indicating a clear domain gap compared to PASCAL VOC dataset. In the SF-FSDA cross-domain experiments, 5 unlabeled images are regarded as the few-shot target domain for training, and the test set containing cat and car objects is adopted for the model performance evaluation on SF-FSDA.

\noindent\textbf{Training Details.} \emph{Image Synthesis}: The data factory is based on the StyleGAN2 structure and initialized with the publicly available cat and car image synthesis pretrained weights in~\cite{karras2020analyzing}. \emph{Label Synthesis}: As the minimum human effort, we manually label 10 synthesized images. \emph{Source Pretraining and Target Adaptation}: The object detection model in the source pretraining and target adaptation stage is based on the Single Shot MultiBox Detector (SSD)~\cite{liu2016ssd} model. We synthesize 200 and 250 samples in total in the in-domain and cross-domain experiments, respectively.

\noindent\textbf{Baseline Setup.} In Table~\ref{table:in_domain} and Table~\ref{table:cross_domain}, the ``Few-Shot FT" represents that the object detection model is fine-tuned on the few-shot manually-labeled images from our data factory. In Table~\ref{table:cross_domain}, the ``CycleGAN", ``MUNIT" and ``CUT" conduct the corresponding image translation methods between the synthesized images from the pretrained StyleGAN2 model and the few-shot target domain images to generate the target domain-like images, and adopt the same annotations generated by our data factory. Oracle performance in Table~\ref{table:in_domain} is reached by training the object detection model on the training set of PASCAL VOC. Oracle performance in Table~\ref{table:cross_domain} is obtained by~\cite{inoue2018cross} for the traditional domain adaptive object detection, which is not few-shot or source-free.

\subsection{Experimental Results} \label{sec:experiment_results}
Through the quantitative and qualitative in-domain and SF-FSDA cross-domain experimental results, we show that our proposed model effectively synthesizes the image samples and the corresponding object bounding box and category labels, and mitigates the source and target domain gap through the guidance of text and few-shot target examples. 

\begin{figure*}
    \centering
    \includegraphics[width=\textwidth]{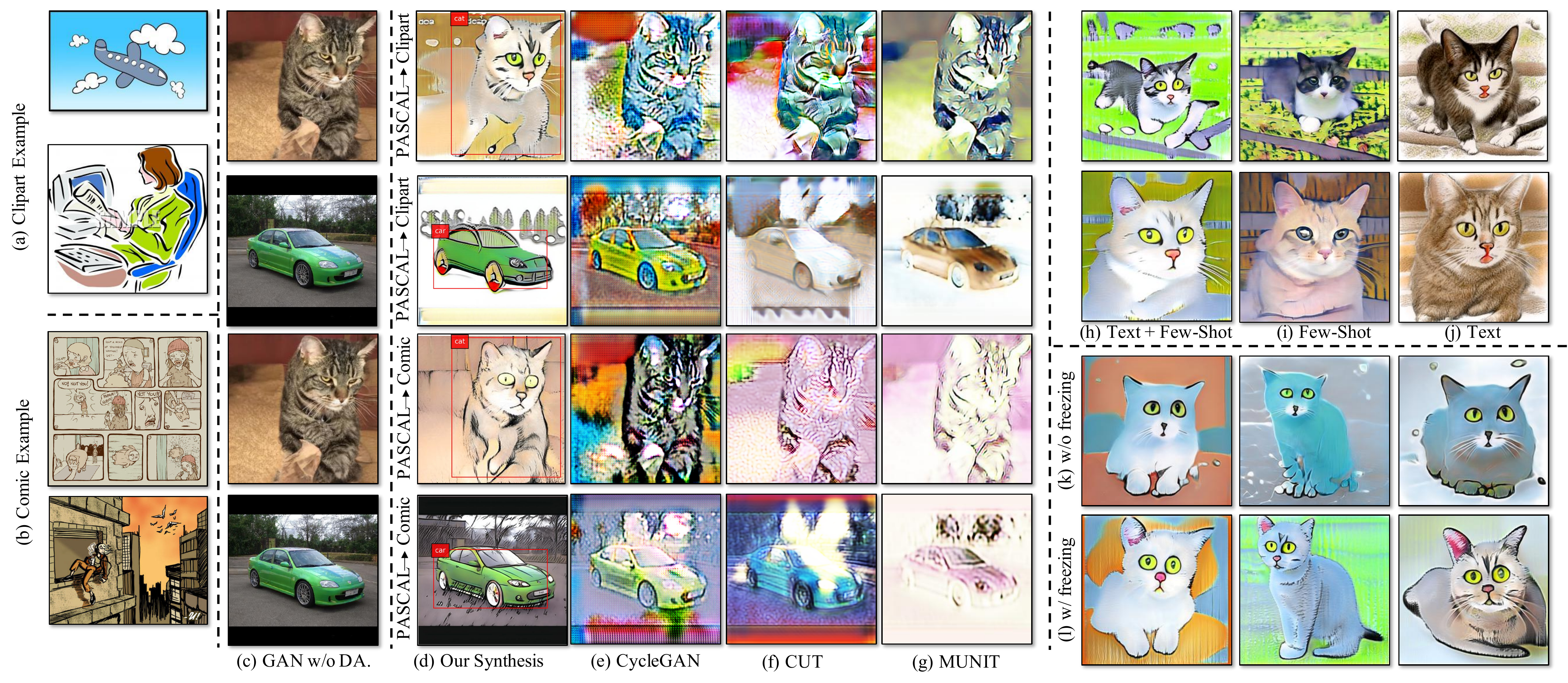}
    % \vspace*{-25pt}
    \vspace{-15pt}
    \caption{(a)-(b) are the exemplar images from the Clipart1k and Comic2k dataset. (c) are the synthesized images from the publicly available pretrained GAN weights, without conducting domain adaptation and used in Table~\ref{table:in_domain}. It is notable that our data factory does not have the requirement of on which style images the GAN model is pretrained, and we just adopt the publicly available pretrained weights provided in~\cite{karras2020analyzing}. (d) are the synthesized image and annotations from our proposed data factory in Table~\ref{table:cross_domain}. (e)-(g) are the results generated by other image translation methods in Table~\ref{table:cross_domain}. (h)-(j) are the ablations for guidance as in Table~\ref{table:ablation_study_guid}. (k)-(l) are the ablations w/ and w/o freezing strategy as in Table~\ref{table:ablation_study_stg}.}
    \label{fig:visual_comp}
    \vspace{-15pt}
\end{figure*}

\noindent\textbf{In-Domain Experiments.} As shown in Table~\ref{table:in_domain} and Fig.~\ref{fig:visual_comp}\textcolor{red}{c}, our synthesized images and corresponding bounding box labels can be used to train the model for the object detection on the same/similar domain, improving the few-shot object detection performance from 50.86\%, 41.57\% to 64.37\%, 52.73\% on the ``Cat" and ``Car" objects detection, respectively. It opens up a new avenue for the few-shot object detection task, by manually labeling object bounding box in a few images, synthesizing enough image samples and bounding box labels automatically with our proposed efficient labeled data factory, and then training the object detection model with the synthesized images and labels. 

\begin{table}[t]
\centering
\caption{In-domain experiments on PASCAL VOC. The results are reported on average precision (AP).}
\vspace*{-5pt}
    \begin{tabular}{c|c|c|c}
		\hline\thickhline
		\rowcolor{mygray}
        &\multicolumn{3}{c}{Method}\\ 
        \cline{2-4}
        \rowcolor{mygray} \multirow{-2}{*}{Classes}& Few-Shot FT& Ours & Oracle\\
        \hline \hline
        Cat & 50.86 & \textbf{64.37} & 86.48\\
        Car & 41.57 & \textbf{52.73} & 72.18\\
        \hline
        \end{tabular}
\label{table:in_domain}
\vspace{-5pt}
\end{table}

\begin{table}[t]
\centering
\caption{Comparison of label generation ways, pseudo label \emph{vs.} our data factory, PASCAL VOC$\rightarrow$ Clipart.}
\vspace*{-5pt}
	\begin{tabular}{c|cc}
		\hline\thickhline
				\rowcolor{mygray}
\multicolumn{3}{c}{Label Generation}\\
\hline
\rowcolor{mygray}Class&Pseudo-Label & Our Label Synthesis\\
\hline \hline
Cat&25.75 & \textbf{32.50}\\
Car&53.52 & \textbf{55.67}\\
\hline
\end{tabular}
\vspace{-10pt}
\label{table:pl_vs_syn}
\end{table}

\noindent\textbf{SF-FSDA Cross-Domain Experiments.} In Table~\ref{table:cross_domain} and Fig.~\ref{fig:visual_comp}\textcolor{red}{d}-\textcolor{red}{g}, the quantitative and qualitative results are shown on the benchmark, PASCAL VOC$\rightarrow$ Clipart, Comic, respectively. Taking the PASCAL VOC$\rightarrow$ Clipart benchmark as the example, compared with the pure source baseline, all of the image style adaptation based methods bring performance improvement, verifying the benefits of the style adaptation based methods for narrowing the domain gap. Among the image style adaptation based methods, it is shown that our proposed data factory based method surpasses other image translation based methods, CycleGAN \cite{zhu2017unpaired}, MUNIT \cite{huang2018multimodal}, and CUT \cite{park2020contrastive}. It proves the advantage of our method for synthesizing the target-domain like image, with the guidance of both the few-shot samples and the text knowledge. Moreover, compared with the few-shot manual annotations, the automatically synthesized annotations can further improve the performance from 30.94\%, 52.97\%  to 32.50\%, 55.67\%. It verifies the effectiveness of the automatically generated images and annotations for the SF-FSDA problem. Similarly, on the PASCAL VOC$\rightarrow$ Comic benchmark, the effectiveness of our proposed method for SF-FSDA is also validated.

\begin{table}%[b]
\centering
\vspace{-10pt}
\caption{SF-FSDA cross-domain experiments.}
\vspace{-5pt}
	\begin{tabular}{c|c|c|c|c|c|c|c}
		\hline\thickhline
				\rowcolor{mygray}
&\multicolumn{7}{c}{Method}\\ 
\cline{2-8}
\rowcolor{mygray} \multirow{-2}{*}{Classes}& Source &Few-Shot FT & CycleGAN & MUNIT & CUT & Ours & Oracle\\
\hline \hline
\multicolumn{8}{c}{PASCAL VOC $\rightarrow$ Clipart} \\
\hline
Cat & 17.25 & 30.94 & 27.01 & 21.46 & 24.57 & \textbf{32.50} & 35.07 \\
Car & 43.04 & 52.97 & 55.11 & 54.62 & 54.72 & \textbf{55.67} & 57.38 \\
\hline
\multicolumn{8}{c}{PASCAL VOC $\rightarrow$ Comic} \\
\hline
Cat & 16.36 & 33.01 & 23.51 & 37.28 & 36.81 & \textbf{37.74} & 39.99 \\
Car & 39.02 & 51.05 & 42.20 & 41.31 & 46.68 & \textbf{54.68} & 52.76 \\
\hline
\end{tabular}
% \vspace{-15pt}
\label{table:cross_domain}
\end{table}

\begin{table}[t]
\caption{Ablation study, PASCAL VOC $\rightarrow$ Clipart (Cat).}
\vspace{-5pt}
\begin{subtable}[h]{0.47\textwidth}
	\centering
% 	\vspace{-10pt}
	\begin{tabular}{ccc|c}
		\hline\thickhline
				\rowcolor{mygray}
\multicolumn{4}{c}{Ablations}\\
\hline
\rowcolor{mygray} Source & Only Few-Shot & Only Text & Few-shot+Text\\
\hline \hline
17.25 & 28.24 & 18.60 & \textbf{32.50}\\ 
\hline
\end{tabular}
% \vspace*{-3pt}
\caption{Ablation study for the text and few-shot images guidance from the target domain, measured with AP performance on Clipart.}
\label{table:ablation_study_guid}
\end{subtable}
\hfill
\begin{subtable}[h]{0.45\textwidth}
\centering
% \vspace{-10pt}
	\begin{tabular}{c|c}
		\hline\thickhline
				\rowcolor{mygray}
\multicolumn{2}{c}{Ablations}\\
\hline
\rowcolor{mygray} w/o freezing & w freezing\\
\hline \hline
0.64 & \textbf{0.68}\\ 
\hline
\end{tabular}
% \vspace*{-3pt}
\caption{Ablation study for freezing strategy during image synthesis training, measured with the LPIPS distance~\cite{ojha2021few}($\uparrow$).}
\label{table:ablation_study_stg}
\end{subtable}
% \vspace*{-10pt}
\vspace{-20pt}
\end{table}

\noindent\textbf{Ablation Study.} In our proposed efficient labeled data factory for SF-FSDA, the style of the generated samples is guided by the few-shot image samples and/or the text guidance. In order to explore the effect of different types of guidance, we compare the performance of different ablations of the full model. From the quantitative comparison in Table~\ref{table:ablation_study_guid}, it is shown that both the few-shot samples and text guidance contribute to the final image synthesis results. From the qualitative results shown in Fig.~\ref{fig:visual_comp}\textcolor{red}{h}-\textcolor{red}{j}, taking the ``comic" style as the example, the text guidance provides the general knowledge on what the ``comic" images look like, while the few-shot images guidance indicates how the ``comic" images are on the target domain. Moreover, the text knowledge from the target domain prevents overfitting to the few-shot samples. On the other hand, it is proven that our model is flexible, still reaching effective synthesis results even when one of the text and few-shot samples guidance is not available. Moreover, the ablation study on the freezing strategy during training is conducted, which is shown in Fig.~\ref{fig:visual_comp}\textcolor{red}{k}-\textcolor{red}{l} and Table~\ref{table:ablation_study_stg}. It is shown that the freezing strategy for the image synthesis training can help prevent overfitting to the few-shot samples in the target domain and preserve the diversity of the image synthesis results.

\textbf{Source Domain Pretraining.} 
To investigate the impact of a source domain pretrained object detection model on SF-FSDA, we conducted a comparison of the SF-FSDA performance with and without the pretrained model for the cat of PASCAL VOC$\rightarrow$Clipart dataset. For the experiment with pretrained model, the pretrained object detection model on source domain is fine-tuned with our generated images and labels on the target domain as done in Sec.~\ref{sec:fact_method}. For the experiment without pretrained model, the object detection model is trained from scratch with our generated images and labels on the target domain. Our results show that the SF-FSDA performance significantly improves with the use of a pretrained model from a large-scale source domain, yielding 32.50\% accuracy with source pretrained compared to 17.74\% without source pretrained. This suggests that leveraging a pretrained model from a relevant large-scale source domain can provide substantial benefits for the success of SF-FSDA.

\begin{figure}[t]
 \centering
 \includegraphics[width=\linewidth]{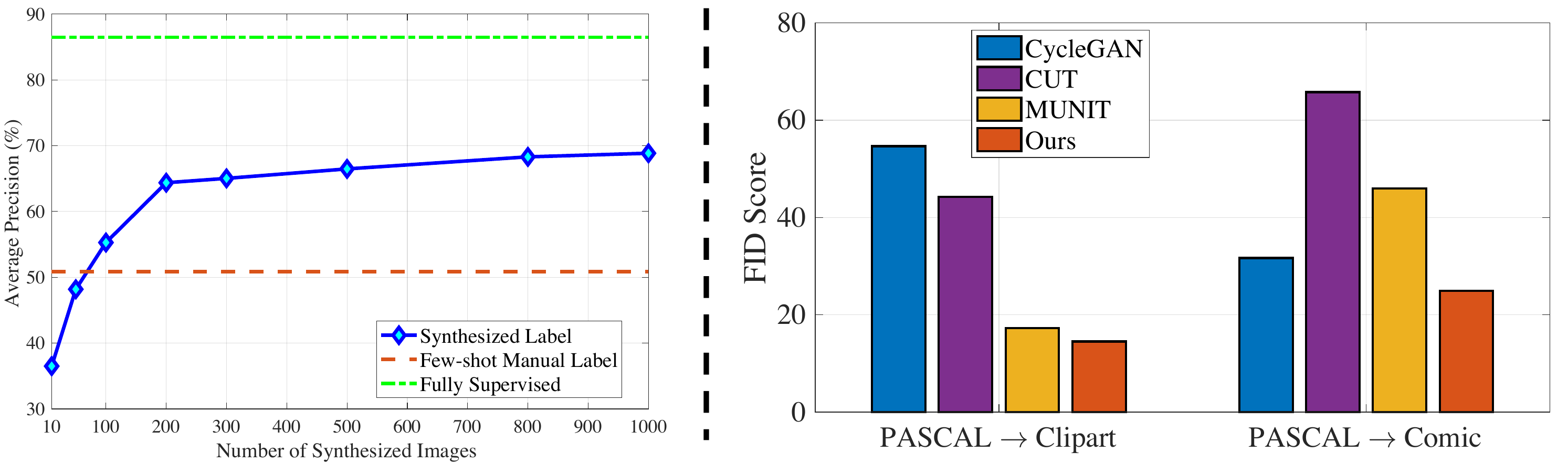}
 % \vspace*{-20pt}
 \vspace{-15pt}
 \caption{Left: Object detection performance with different numbers of synthesized images and annotations, under the setting of Table~\ref{table:in_domain}. Right: Synthesized images quality comparison between our proposed data factory and other image translation based methods, measured with FID score ($\downarrow$).}
 \label{fig:sample_study}
 \vspace{-10pt}
\end{figure}

\noindent\textbf{Number of Synthesized Image Samples Study.} In order to figure out the effect of the number of synthesized images and annotations from the efficient labeled data factory, the object detection performance with different numbers of synthesized images and samples are shown in Fig.~\ref{fig:sample_study}. It is shown that the object detection performance improves as more images and annotations are synthesized.

\noindent\textbf{Pseudo Label \emph{vs.} Our Label Synthesis.} Under the cross-domain experiments setting, an alternative way to our label synthesis through the efficient labeled data factory is to apply the source domain pretrained object detection model on our synthesized images to generate the pseudo-label. In Table~\ref{table:pl_vs_syn}, the pseudo-label and the label synthesis with our efficient labeled data factory ways for label generation are compared. It is shown that our synthesized label with the data factory performs better than the pseudo-label for the SF-FSDA problem. It is because the pseudo-label is noisy and of low quality, resulting from the difference between the source domain and the synthesized images and the source-free condition. In contrast, our efficient labeled data factory synthesizes the annotation with the help of the image synthesis and few-shot manual annotations, bringing high-quality automatic annotations.
    
\begin{figure}[t]
    \centering
    \includegraphics[width=\linewidth]{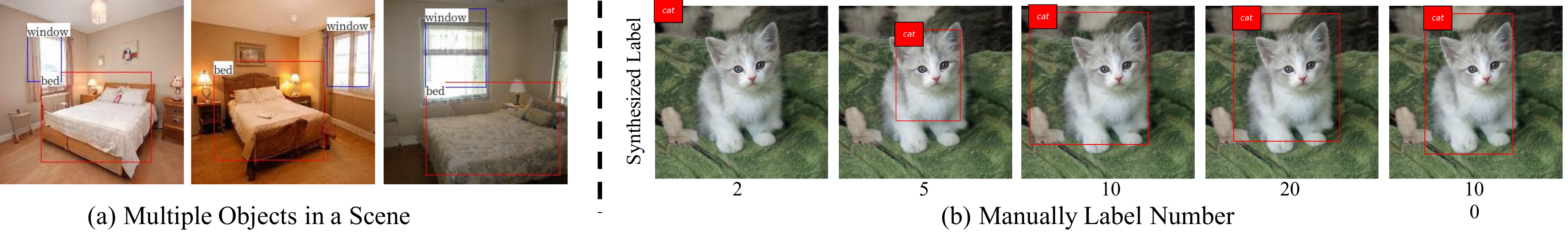}
    \vspace{-20pt}
    \caption{(a) Multiple objects ``window" and ``bed" are synthesized and labeled in a single indoor scene image. (b) Manual label number for training label synthesis branch study. It is observed that increasing manual label number for training label synthesis branch helps generate more precise bounding box and detect more accurate objects.}
    \label{fig:manual_label_num_bedroom}
    \vspace{-10pt}
\end{figure}

\textbf{Multiple Objects in a Single Image.}
In order to further prove the effectiveness of our proposed method for the scenario that there are multiple objects in a single image, we conduct the experiment where the indoor scene images are synthesized and the corresponding ``window" and ``bed" objects label are generated. The qualitative experimental results are shown in Fig.~\ref{fig:manual_label_num_bedroom}. It is shown that our method can effectively synthesize and label the image where multiple objects appear in a single scene.

\textbf{Manual Label Number Study.}
In order to study the influence of the manual label number for training label synthesis branch, we change the manual label number and train the efficient labeled data factory on the cat category, whose label synthesized results are shown in Fig.~\ref{fig:manual_label_num_bedroom}. It is proven that increasing manual label number for training label synthesis branch helps generate more precise bounding box and detect more accurate objects.

\section{Conclusion}
We propose and tackle the SF-FSDA problem, which studies the domain adaptive object detection problem under \emph{source-free} and \emph{few-shot} conditions. In order to overcome the problem, we present a new efficient labeled data factory based method, which can synthesize the infinite target domain-like images and corresponding annotations without relying on the source domain. The image synthesis branch is guided by the few-shot image samples and text from the target domain, and the image annotation branch only requires the minimum human effort (\ie, few-shot manual labels) to generalize the label to the rest of the synthesized images. The synthesized target domain-like images and annotations are further utilized to fine-tune the source domain pretrained object detection model. The proposed approach is validated in various settings and surpasses other methods, demonstrating its effectiveness for SF-FSDA problem.

\bibliography{collas2023_conference}
\bibliographystyle{collas2023_conference}

\appendix
\section*{\textbf{Appendix}}
In this supplementary material, we provide the additional information for,
\begin{itemize}
    \item[\textbf{A}] detailed framework implementation and training parameters,
    \item[\textbf{B}] detailed information about datasets involved in the experiments,
    \item[\textbf{C}] additional qualitative experimental results,
    \item[\textbf{D}] additional quantitative experimental results.
\end{itemize}

\section{Framework Implementation}
This section provides detailed implementation of the image synthesis and label synthesis branches of our proposed efficient labeled data factory, and their corresponding training details. 

\subsection{Image Synthesis Branch}

\noindent\textbf{Network Structure.}
The network is built upon the structure of StyleGAN2~\cite{karras2020analyzing},
with additional dual-generator and dual-discriminator design to incorporate extra text guidance and relaxed realism strategy for avoiding overfitting to few-shot image samples from the target domain.

To bring in the text guidance~\cite{gal2021stylegannada}, the dual-generator strategy is deployed. We have two generators, $G_{t}$ to be adapted, and a frozen $G$ for image changing reference. The second generator $G$ follows the same structure as the original one, and remains frozen during the whole training process to keep generating original images without adaptation.

In order to prevent the synthesized images from overfitting to the few-shot image samples of the target domain, the dual-discriminator strategy is deployed, introducing the relaxed realism~\cite{ojha2021few}.  As introduced in Sec.~\textcolor{red}{3.2} of the main paper, the idea is to only discriminate the fake and real images on the image level with the discriminator $D_{img}$ when the corresponding latent code $z$ is sampled from a small anchor region. The image-level discriminator $D_{img}$ follows the original design as in~\cite{karras2020analyzing}. 
When $z$ is sampled outside of the region, we compute the discriminator loss only on the patch level with the second discriminator $D_{patch}$, which is the subset of the discriminator $D_{img}$. The two discriminators are trained at a designed frequency for the purpose of preserving image diversity while still leveraging whole-image guidance. The process is controlled with the sampling frequency hyper-parameter $\lambda_f$, which indicates the frequency of sampling from the anchor region and computing the image-level loss instead of the patch-level loss.

\noindent\textbf{Training Parameters.}
We adjust the weight of image guidance $\lambda_1$ and the weight of text guidance $\lambda_2$ in Eq. (\textcolor{red}{5}) of the main paper under different scenarios, to balance different guidances. Under the PASCAL VOC$\rightarrow$Clipart setting, we set both weights to 1.0. Under the PASCAL VOC$\rightarrow$Comic setting, we increase the weight of text guidance to 5.0. 
Another parameter is the sampling frequency from the randomly sampled small anchor region, $\lambda_f$, which decides how often we compute the discriminator loss on the whole image level. We set the frequency to 2, alternatively computing the loss on the image level and the patch level. The rest training details follow the StyleGAN2~\cite{karras2020analyzing} with the augmentation strategy introduced in~\cite{karras2020training}. The training iteration for image synthesis is set as 1000.

In order to avoid overfitting to the few-shot image samples from the target domain, we develop the freezing strategy for image synthesis training (see the \emph{image synthesis training strategy} part in Sec.~\textcolor{red}{3.2} of the main paper). More specifically, we adapt the generator $G_{t}$ only on specific feature layers while freezing the rest part of the network. For all the experiments, we only update the weights of intermediate layers from the third to the last one. 
Accordingly, we freeze the image-level and the patch-level discriminator, $D_{img}$ and $D_{patch}$, except for the final layer.

A latent code $\z$ is randomly sampled for each image with a dimension of 512 for detection labeling. In order not to generate low quality images, the latent code is truncated by the average latent code to avoid sampling from low probability density region. More details can be found in Sec. B ``Truncation trick in W" of \cite{karras2019style}.

\subsection{Label Synthesis Branch}

\noindent\textbf{Network Structure.}
Our label synthesis branch is built by utilizing StyleGAN2~\cite{karras2020analyzing} generator acquired in the image synthesis step as the backbone network, and then adding different prediction heads on this basis.
For the backbone network, we take the intermediate feature map with the resolutions (4, 8, 16, 32, 64) considering the memory consumption. Then we upsample those feature maps with bilinear interpolation to the resolution of 128, and concatenate them together to feed forward to the 
three prediction heads for keypoint, offset, and bounding box size predictions, respectively. The prediction head is composed of, 3$\times$3 convolutional layer, ReLU, and 1$\times$1 convolutional layer.

\noindent\textbf{Training Parameters.}
We mainly follow the training parameters and details in~\cite{zhou2019objects}.
We set the hyper-parameters in Eq. (\textcolor{red}{7}) to $\lambda_{off}=1.0$ and $\lambda_{size}=0.5$. We adopt the SGD optimizer for training, with the learning rate as 0.0001 and the weight decay as 0.0001. Keypoints are predicted on a heatmap with the resolution of 128. The training iteration for label synthesis is set as 1000.

\section{Datasets Information}
In Sec.~\textcolor{red}{4.1} of the main paper, we provide the information about the datasets which are involved in our experiments. We here further provide additional datasets information.

\noindent\textbf{PASCAL VOC.} PASCAL VOC 2007 \& 2012 dataset~\cite{everingham2010pascal} contains natural images. Each image in PASCAL VOC dataset includes the object class, pixel-level semantic label, and object bounding box annotations, serving as an important benchmark for the image classification, semantic segmentation and object detection tasks. Our experiment is related to the object detection task on PASCAL VOC dataset. PASCAL VOC dataset customizes the license, especially the images collected from the Flickr website, \ie, PASCAL VOC dataset grants the limited, non-transferable, non-sublicensable, revocable license to access and use the data.

\noindent\textbf{Clipart1k.} Clipart1k dataset includes the clipart images collected from the CMPlaces dataset~\cite{castrejon2016learning} and two image search engines~\cite{inoue2018cross}. Clipart1k is meant for the education and research purposes only. 

\noindent\textbf{Comic2k.} Comic2k dataset covers the comic images collected from BAM!~\cite{wilber2017bam}. Comic2k dataset is meant for the education and research purposes only.

\noindent\textbf{Watercolor2k.} Watercolor2k dataset contains the watercolor images collected from BAM!~\cite{wilber2017bam}, which are meant for the education and research purposes only.

\section{Additional Qualitative Results}
\subsection{Freezing Strategy for One-Shot Target Domain}
\begin{figure*}[t]
    \centering
    \includegraphics[width=\linewidth]{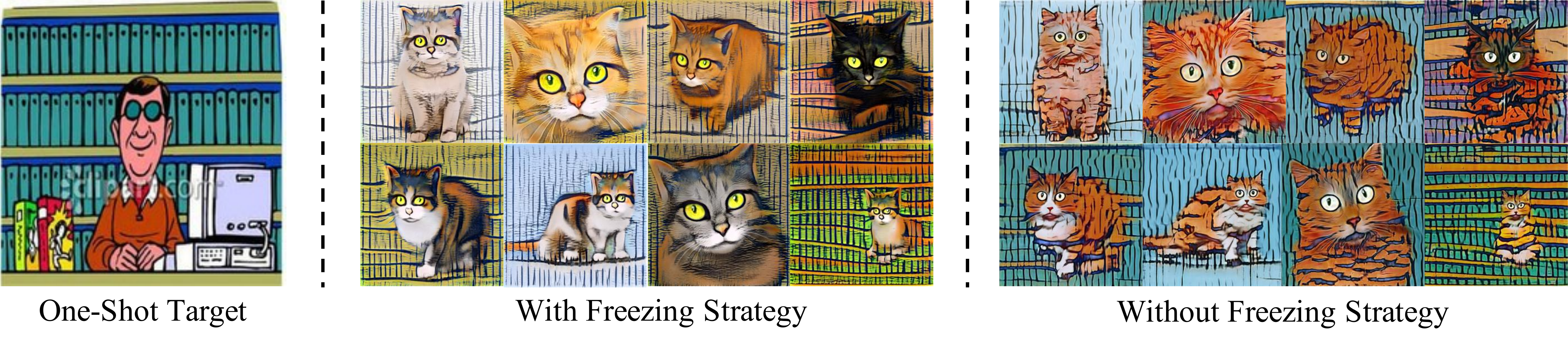}
    \caption{Qualitative results comparison, with/without freezing strategy for image synthesis training, under the one-shot target domain condition. It is shown that the freezing strategy can help to improve the image generation diversity and to prevent overfitting to the one-shot target domain effectively.}
    \label{fig:one_shot}
\end{figure*}
In Fig.~\textcolor{red}{5} (k)-(l) of the main paper, we show the qualitative comparison results for the ablation study with/without the freezing strategy of image synthesis training. In order to further prove the validity of the freezing strategy under the extreme case, we here provide the qualitative comparison in Fig.~\ref{fig:one_shot} under the one-shot target domain condition, \ie, there is only one image available on the target domain. From Fig.~\ref{fig:one_shot}, it is shown that the freezing strategy is especially important for improving the image generation diversity and preventing overfitting to the one-shot image samples under the challenging one-shot condition.

\subsection{Additional Image Synthesis Results on Human Face}
\begin{figure*}[t]
    \centering
    \includegraphics[width=\linewidth]{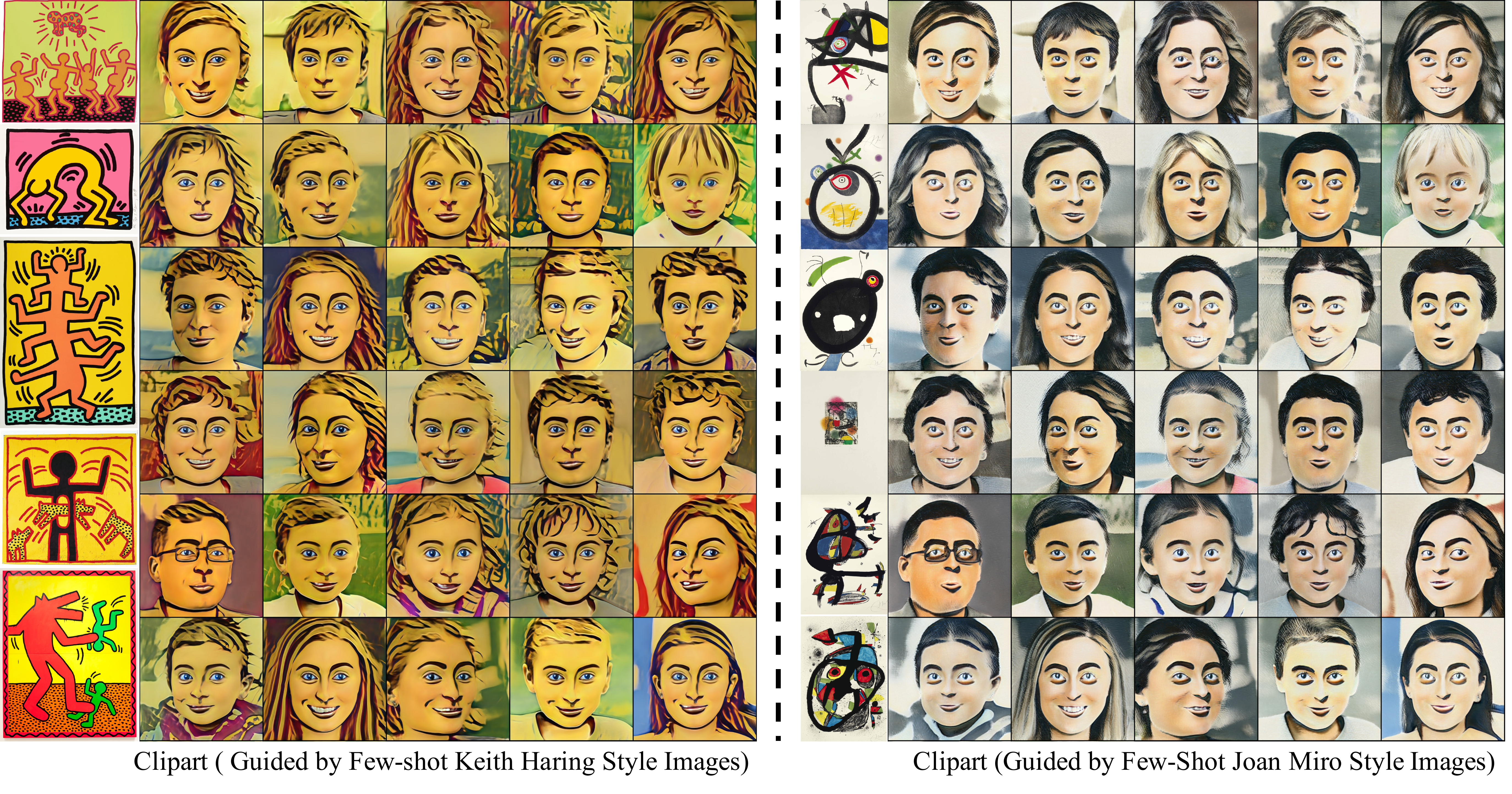}
    \caption{Image synthesis results on human face. The image synthesis is guided by the text guidance ``clipart" and the few-shot image samples guidance from ``Keith Harring" and ``Joan Miro" style paintings. The first column of the left part and the right part is the example of the ``Keith Harring" and ``Joan Miro" painting style.}
    \label{fig:ffhq}
\end{figure*}
In Sec.~\textcolor{red}{4} of the main paper, we provide the experimental results for SF-FSDA under the PASCAL VOC$\rightarrow$Clipart, Comic settings, to synthesize the cat and car objects. In order to further prove the ability of our data factory for image synthesis, we utilize our proposed data factory to synthesize other objects, human face, guided by the few-shot images and the text. The few-shot image guidance includes different artistic style paintings and the provided text guidance is ``clipart". The qualitative results shown in Fig.~\ref{fig:ffhq} prove that our proposed data factory effectively synthesizes the target-domain like images under the text and the few-shot image samples guidance.

\subsection{Additional Image Synthesis Results on Adverse Weather Simulation}
\begin{figure*}[t]
    \centering
    \includegraphics[width=\linewidth]{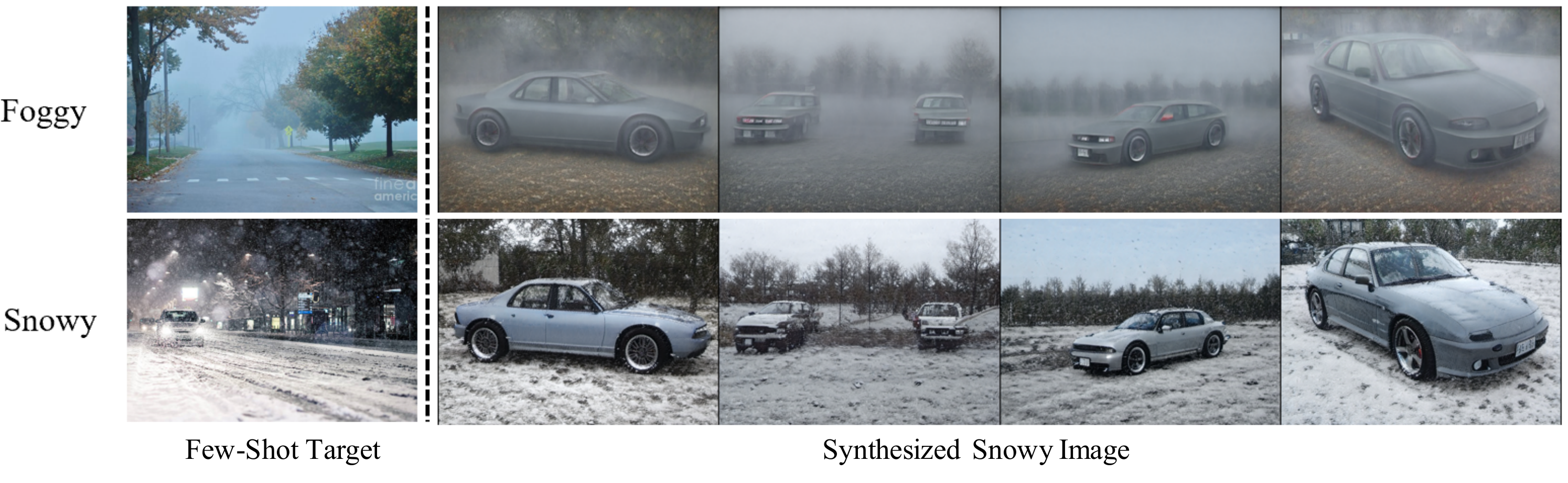}
    \caption{Image synthesis results on car, with few-shot image and text guidance of ``snowy" and ``foggy".}
    \label{fig:weather}
\end{figure*}

In order to show the possible application of our method to different scenarios, \eg, autonomous driving, we conduct experiments adapting original images of the car category to reflect adverse weathers, \eg, snowy weather. Fig.~\textcolor{red}{3} shows the adapted synthesized images under adverse
weather conditions. For the adaptation, text description from the ``sunny" to
the ``snowy" and ``foggy" together with 5 style images of street scenes under ``snowy" and ``foggy" weather are provided as guidance. As shown in the Fig.~\ref{fig:weather}, our data factory is effective for synthesizing the adverse weather images.

\subsection{Qualitative Results on the Target Domain}
\begin{figure*}[t]
    \centering
    \includegraphics[width=\linewidth]{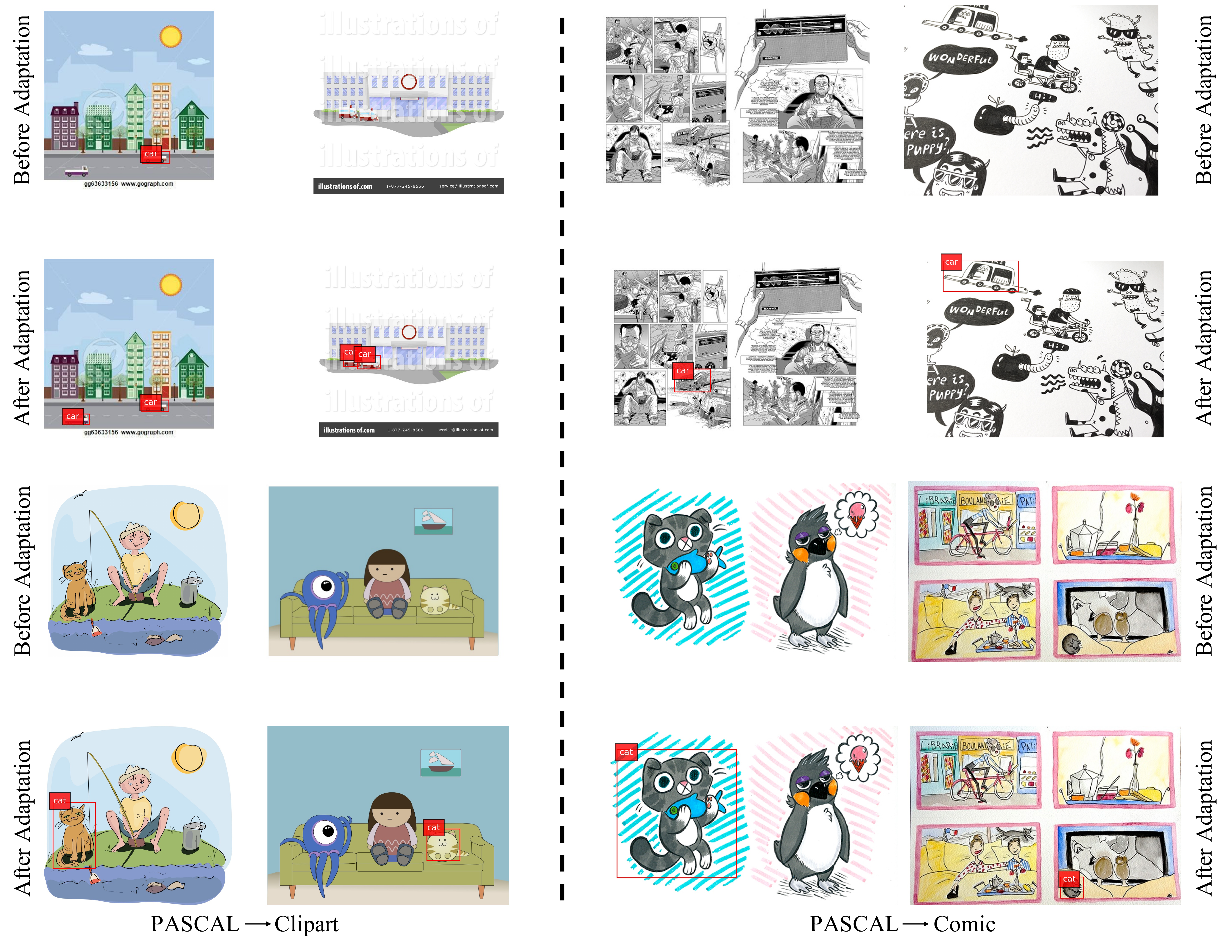}
    \caption{Qualitative object detection results on the target domain, Clipart and Comic. ``Before Adaptation" represents the object detection results when applying the source-pretrained object detection model to the target domain. ``After Adaptation" shows the object detection results after fine-tuning the source-pretrained model on the synthesized images and labels from our proposed data factory. The image without detected bounding box indicates that the model cannot detect the objects in the image.}
    \label{fig:target_detection}
\end{figure*}
In Sec.~\textcolor{red}{4} of the main paper, we show the quantitative robust object detection results on the Clipart1k and Comic2k datasets. In order to further prove the effectiveness of our proposed data factory for robust object detection, we here show the qualitative object detection results on the target domain, \ie, Clipart1k and Comic2k. From the qualitative results in Fig.~\ref{fig:target_detection}, it is shown that our proposed efficient labeled data factory adapts the source pretrained object detection model to the target domain, and improves the object detection performance on the target domain effectively.

\subsection{Additional Baseline Comparison Results}
\begin{figure*}[t]
    \centering
    \includegraphics[width=\linewidth]{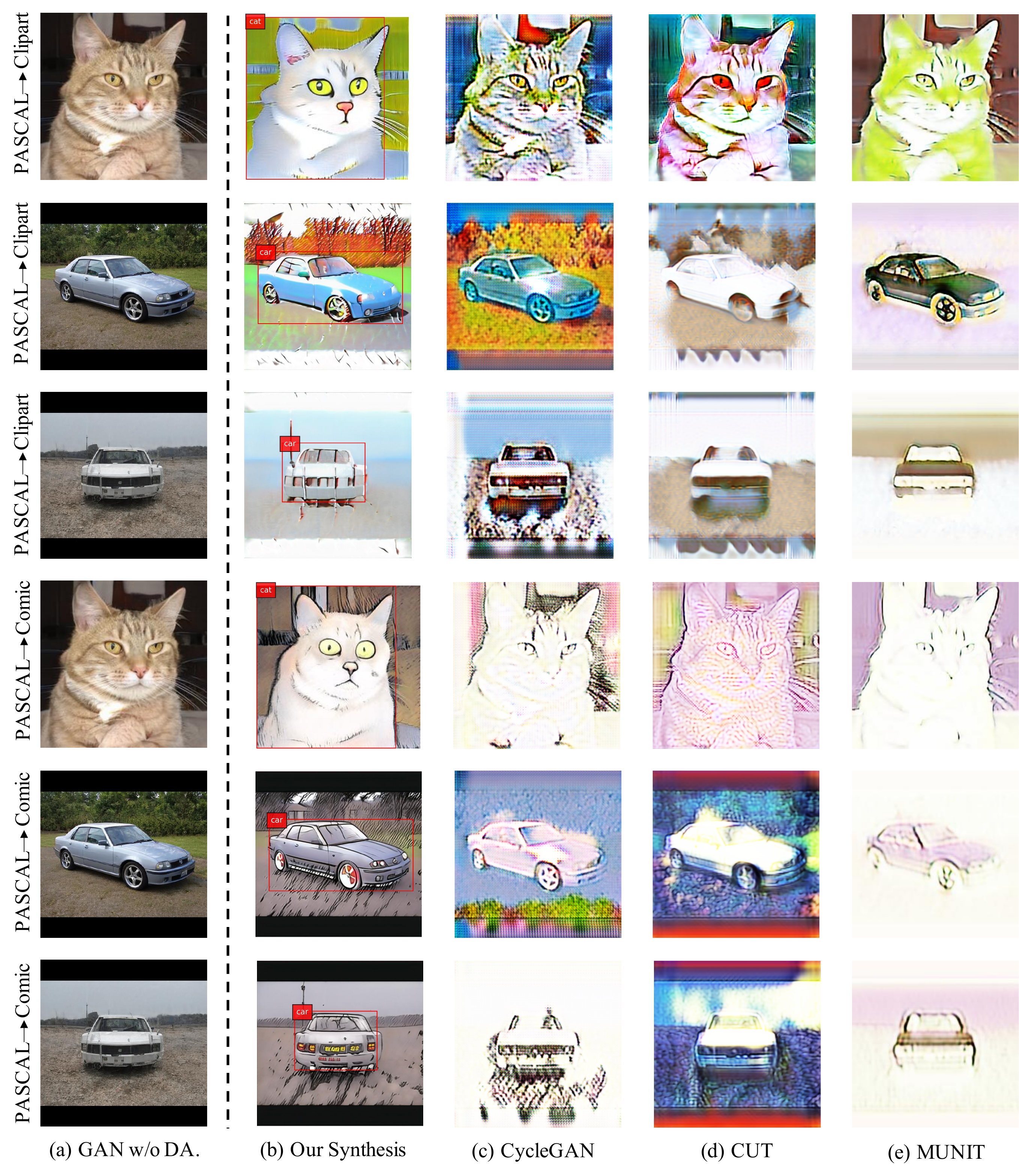}
    \caption{Comparison between data factory and other image translation based methods. (b) is synthesized from our proposed efficient labeled data factory, while (c)-(e) are generated from the image-translation based methods, CycleGAN, CUT and MUNIT.}
    \label{fig:baseline_comp}
\end{figure*}
In Fig.~\textcolor{red}{5}(d)-(g) of the main paper, we show the qualitative comparison results between our proposed efficient labeled data factory method and other image translation based methods, CycleGAN~\cite{zhu2017unpaired}, CUT~\cite{park2020contrastive}, and MUNIT~\cite{huang2018multimodal}. We here provide additional comparison results in Fig.~\ref{fig:baseline_comp} to validate that our proposed efficient labeled data factory can synthesize the images and corresponding object bounding box and category annotations effectively.

\subsection{SF-FSDA: PASCAL VOC$\rightarrow$Watercolor}
\begin{figure*}[t]
    \centering
    \includegraphics[width=\linewidth]{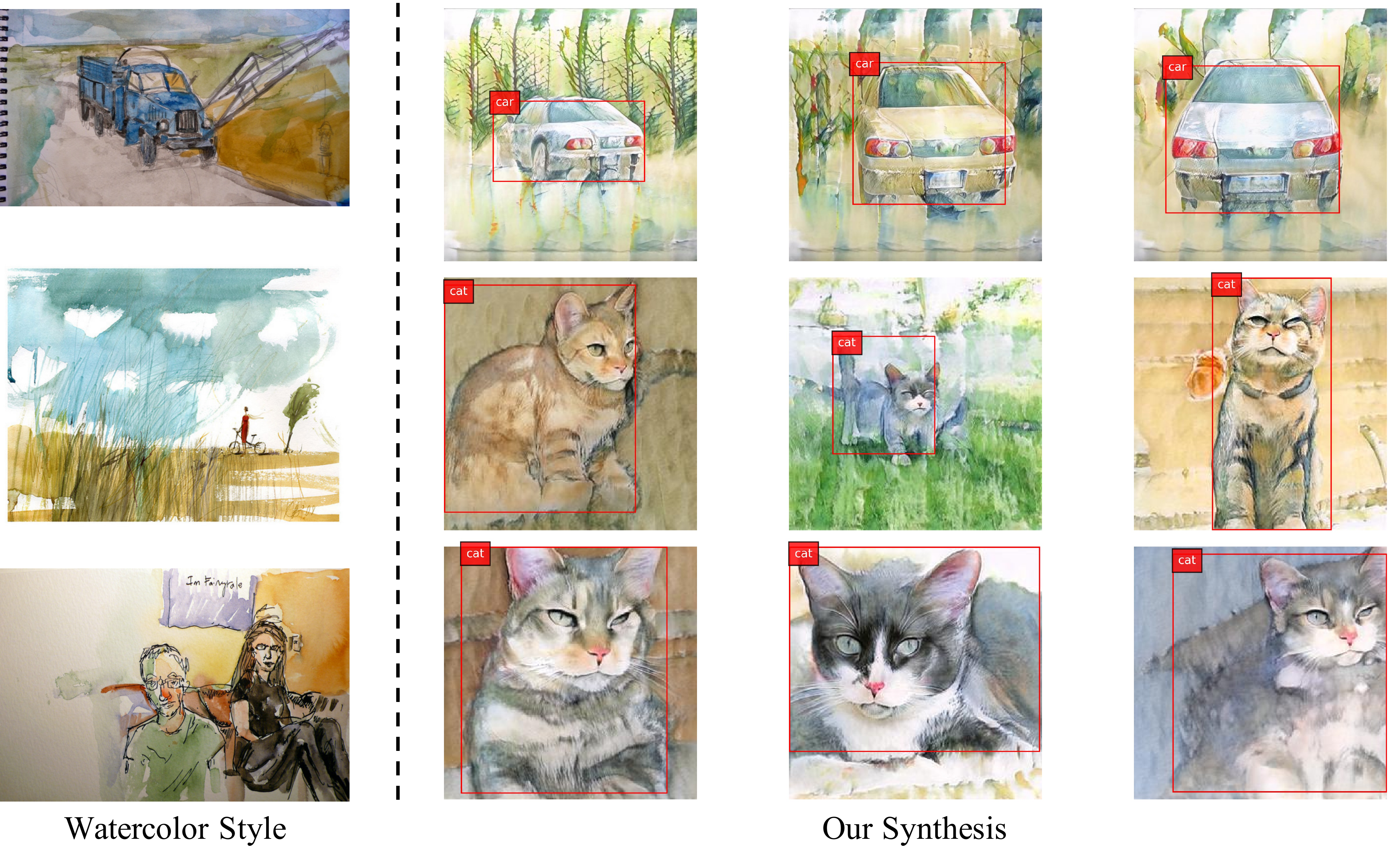}
    \caption{Image and label synthesis results for SF-FSDA, under the PASCAL VOC$\rightarrow$Watercolor setting. The text guidance is ``watercolor", and the few-shot image samples guidance is the watercolor image from~\cite{inoue2018cross}.}
    \label{fig:watercolor}
\end{figure*}
In Sec.~\textcolor{red}{4} of the main paper, we show quantitative and qualitative experimental results for SF-FSDA under the PASCAL VOC$\rightarrow$Clipart and Comic settings. In order to further verify the validity of our proposed efficient labeled data factory method for SF-FSDA, we conduct the additional SF-FSDA experiments under the PASCAL VOC$\rightarrow$Watercolor setting. Compared to the few-shot fine-tuning baseline as done in Table~\textcolor{red}{3} and Table~\textcolor{red}{4} of the main paper, our proposed efficient labeled data factory based method further improves the object detection performance from 49.03\%, 64.47\% to 52.06\%, 65.56\% for cat and car categories, respectively. In Fig.~\ref{fig:watercolor}, we show the image and label synthesis results from our data factory, under the PASCAL VOC$\rightarrow$Watercolor setting.

\subsection{Comparison to StyleGAN-NADA}
\begin{figure}[t]
    \centering
    \includegraphics[width=\linewidth]{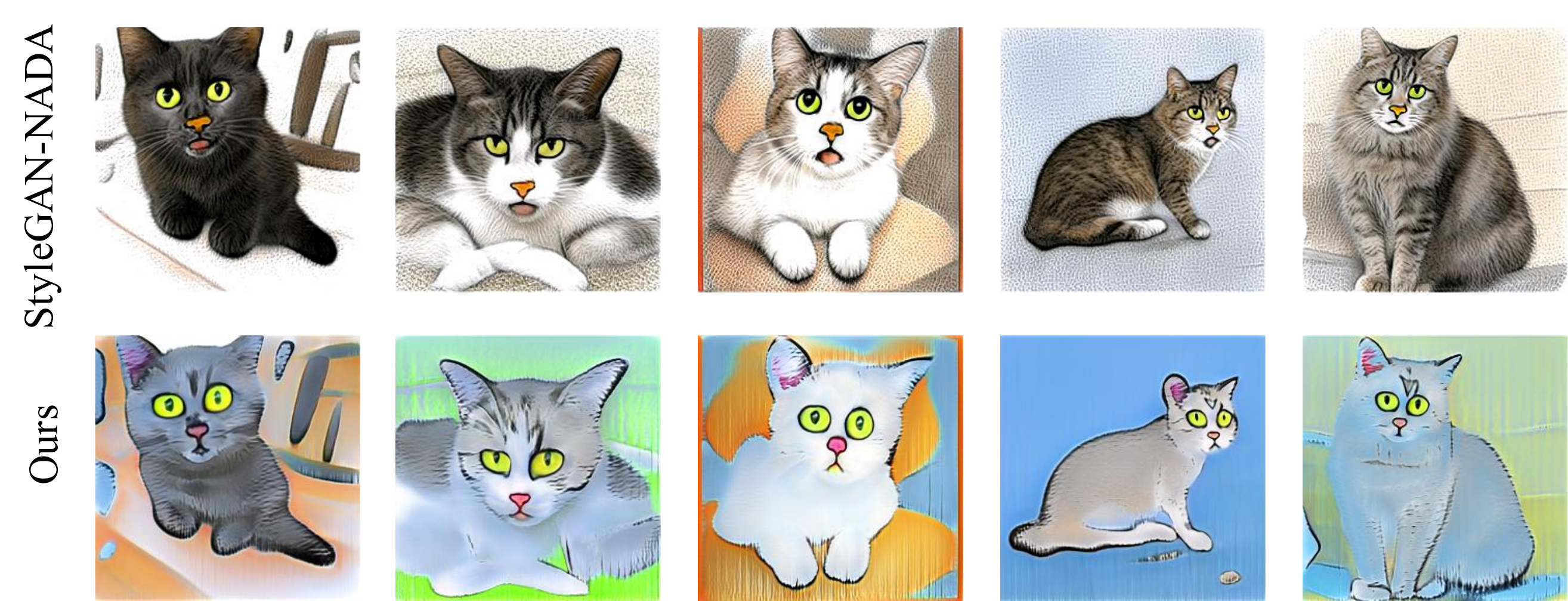}
    \caption{Comparison with StyleGAN-NADA~\cite{gal2021stylegannada} for Clipart image synthesis.}
    \label{fig:stylegannada}
\end{figure}

In order to further validate the effectiveness of our proposed method, we compare our method with recent language guided image synthesis method, StyleGAN-NADA~\cite{gal2021stylegannada}. As shown in Fig.~\ref{fig:stylegannada}, it is observed that the images synthesized by our method are more like Clipart image than StyleGAN-NADA, benefiting from the additional few-shot samples guidance and the freezing training strategy for image synthesis.

\subsection{target domain image guidance samples number study}
\begin{figure}[t]
    \centering
    \includegraphics[width=0.6\linewidth]{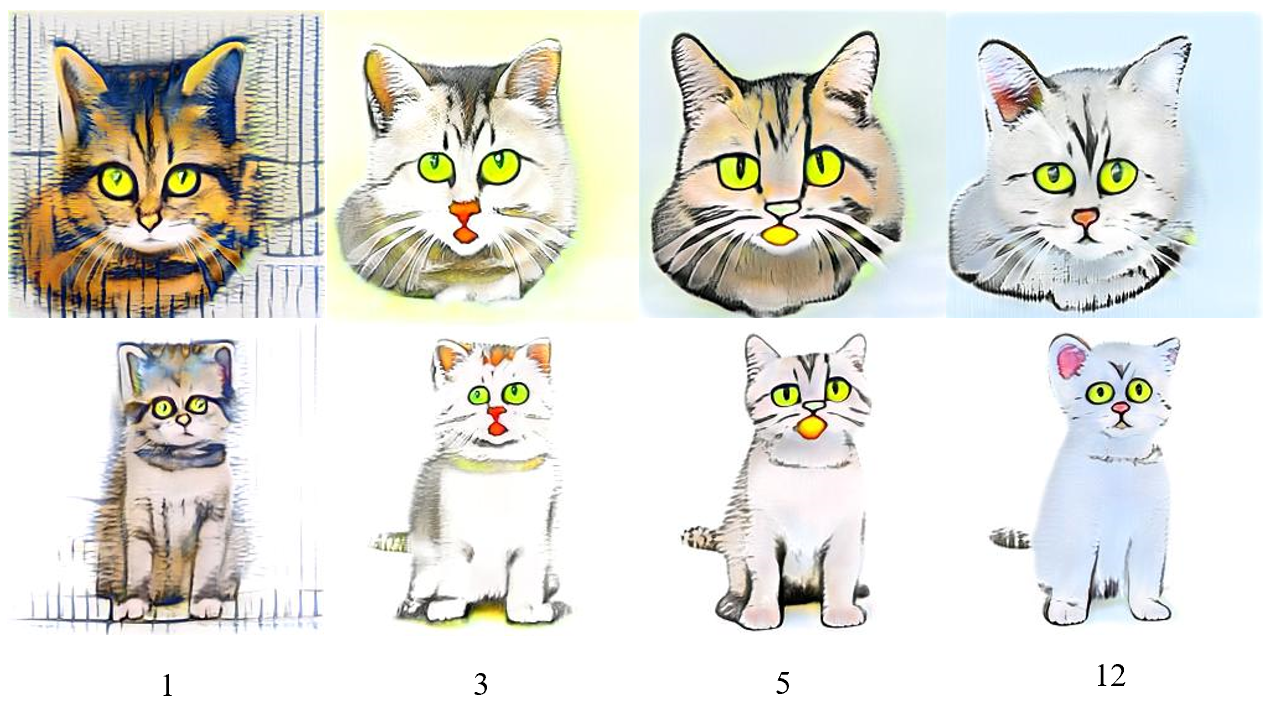}
    \caption{Image synthesis results guided by different number of image guidance samples (1, 3, 5, 12) from the target domain (Clipart).}
    \label{fig:shotnum}
\end{figure}

To explore the effect of target samples number for the image synthesis branch, we compare the image synthesis results guided by varying numbers of target samples for training. Our experimental findings, as illustrated in Fig.~\ref{fig:shotnum}, indicate that our method is able to avoid collapse even with one-shot guidance. However, we observe that using too few examples results in a lack of intra-style differences, ultimately resulting in monochromatic images with a uniform background.

\subsection{freezing strategy study}
\begin{figure}[t]
    \centering
    \includegraphics[width=0.6\linewidth]{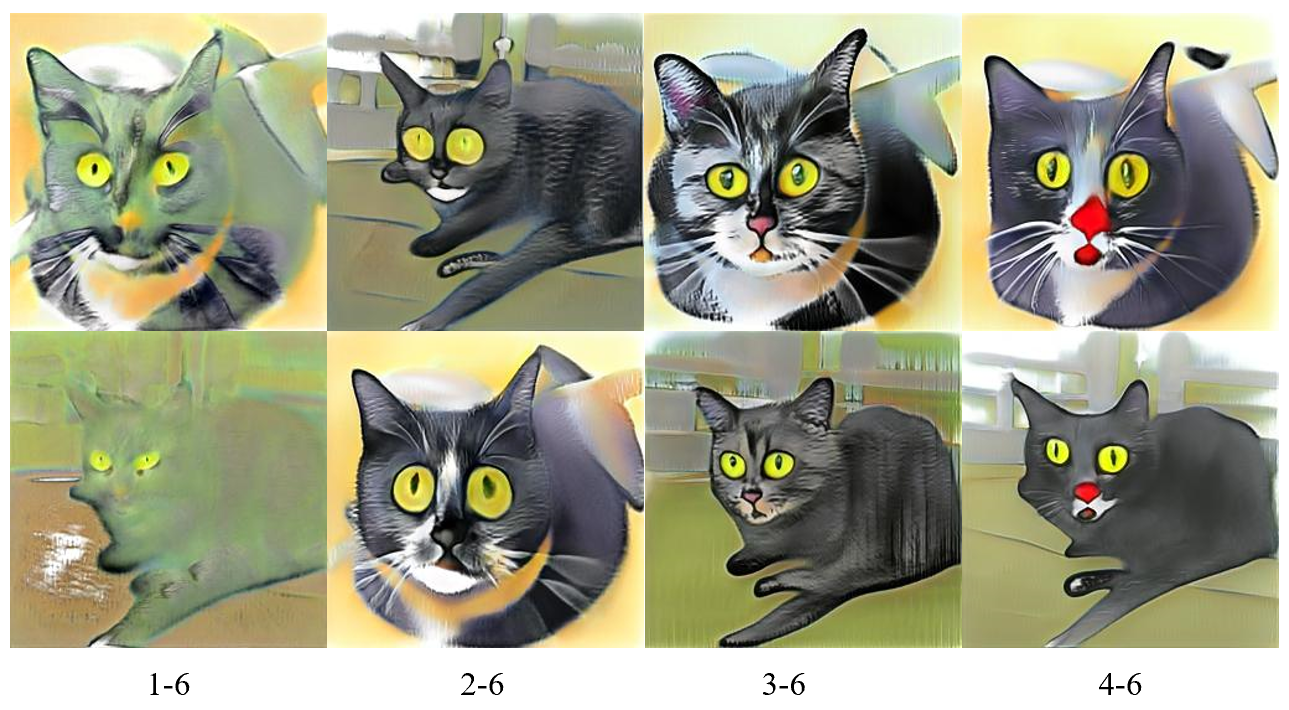}
    \caption{Image synthesis results obtained by fine-tuning different layers of the image generator. Specifically, we denote unfreezing and fine-tuning the first to the last layer of the image generator as 1-6, respectively.}
    \label{fig:glayer}
\end{figure}

We investigate the freezing strategy hyperparameter choice by experimenting with different sets of frozen parameters. We conduct experiments by unfreezing different numbers of layers of the generator and report our findings in Fig.~\ref{fig:glayer}. Specifically, we observe that unfreezing too many layers of the image generator leads to corrupted and blurred generation results, while freezing too many layers limits the model's flexibility for adaptation, resulting in the collapse to monochromatic local patterns. We also experiment with different freezing strategies for the discriminator and find that unfreezing more layers often leads to imbalanced training. Therefore, for all our experiments, we only unfreeze the last linear layer of the discriminator.

\section{Additional Quantitative Results}
\subsection{Comparison to Few-Shot Labeled Target Domain Images}
In order to further prove the effectiveness of our proposed approach, we compare our approach to the few-shot labeled target domain images, where the few-shot target domain images are labeled and then are used to fine-tune the source-pretrained object detection model. The number of manually labeled few-shot target domain images is the same as that of manually labeled synthesized images used for training label synthesis branch in the efficient labeled data factory. Under the PASCAL VOC $\rightarrow$ Clipart ``cat" category setting in Table 3 of the main paper, the few-shot labeled target domain images fine-tuning baseline reaches the performance 29.97\%, while our approach reaches 32.50\%. It is because our efficient labeled data factory can synthesize a large number of images and corresponding label, while the few-shot labeled target domain samples are insufficient.

\subsection{Target Domain Samples Number Study}
As observed in Fig.~\textcolor{red}{S1} of the supp. and Fig.~\textcolor{red}{3(d)} of the main paper, more target samples can improve the diversity of the synthesized image samples and prevent overfitting. Quantitatively, corresponding to Fig.~\textcolor{red}{3(d)}, the object detection performance under PASCAL VOC $\rightarrow$ Clipart ``cat" category is 32.50\%, outperforming the performance corresponding to Fig.~\textcolor{red}{S1}, 23.15\%.

\subsection{Multi-domain study}
To evaluate the effectiveness of our method in multi-target domain scenarios, we conduct experiments on multiple target domains and report the results in Table ~\ref{table:multi_domain}. Our efficient labeled data factory achieves reliable performance on this task. Specifically, as depicted in the table, we observe that transferring from one target domain to another can even improve the performance compared to direct adaptation from the original source pretrained model. This highlights the potential of our method to be used in real-world scenarios with multiple target domains.

\begin{table}[t]
\centering
    \caption{Multi-target domain SF-FSDA results. We evaluate our SF-FSDA model's performance on two adaptation scenarios: PASCAL VOC $\rightarrow$ Comic and PASCAL VOC $\rightarrow$ Clipart $\rightarrow$ Comic. In the first scenario, we adapt the PASCAL VOC trained model on Comic-style training samples synthesized by our data factory. In the second scenario, we first adapt the PASCAL VOC trained model to Clipart-style training samples synthesized by our data factory, and then further adapt it to the Comic-style training samples synthesized by our data factory.}
\vspace*{-5pt}
	\begin{tabular}{c|cc}
		\hline\thickhline
				\rowcolor{mygray}
\rowcolor{mygray}Class&PASCAL VOC $\rightarrow$ Comic & PASCAL VOC $\rightarrow$ Clipart $\rightarrow$ Comic\\
\hline \hline
Cat&37.74 & \textbf{40.14}\\
Car&54.68 & \textbf{57.02}\\
\hline
\end{tabular}
% \vspace{-10pt}
\label{table:multi_domain}
\end{table}

\subsection{Manual label samples number study}
We investigated the impact of the manual labels number on the SF-FSDA performance, whose results are presented in Table~\ref{table:manual_num}. The experimental results reveal that the SF-FSDA performance improves with an increase in the number of manual labels. However, the rate of improvement decreases as the number of manual labels becomes larger. After balancing stability and efficiency, we selected the number of manual labels as 10.

\begin{table}[t]
\centering
\caption{SF-FSDA performance on the Comic target domain for Cat category, with varying numbers of manually labeled samples utilized for label synthesis.}
\vspace*{-5pt}
	\begin{tabular}{ccccc}
		\hline\thickhline
				\rowcolor{mygray}
\multicolumn{5}{c}{Number of Manual Labels}\\
\hline
\rowcolor{mygray}5&10&20&50&100\\
\hline \hline
37.04 & 37.74 & 38.54 & 39.25 & 39.87\\
\hline
\end{tabular}
\vspace{-10pt}
\label{table:manual_num}
\end{table}

\end{document}